# Towards social embodied cobots: The integration of an industrial cobot with a social virtual agent

Matteo Lavit Nicora[1,2], Sebastian Beyrodt[3],
Dimitra Tsovaltzi[3], Fabrizio Nunnari[3], Patrick Gebhard[3], Matteo Malosio[1]

*Abstract*— The integration of the physical capabilities of an industrial collaborative robot with a social virtual character may represent a viable solution to enhance the workers' perception of the system as an embodied social entity and increase social engagement and well-being at the workplace. An online study was setup using prerecorded video interactions in order to pilot potential advantages of different embodied configurations of the cobot-avatar system in terms of perceptions of Social Presence, cobot-avatar Unity and Social Role of the system, and explore the relation of these. In particular, two different configurations were explored and compared: the virtual character was displayed either on a tablet strapped onto the base of the cobot or on a large TV screen positioned at the back of the workcell. The results imply that participants showed no clear preference based on the constructs, and both configurations fulfill these basic criteria. In terms of the relations between the constructs, there were strong correlations between perception of Social Presence, Unity and Social Role (Collegiality). This gives a valuable insight into the role of these constructs in the perception of cobots as embodied social entities, and towards building cobots that support well-being at the workplace.

## I. INTRODUCTION

The industry 4.0 and 5.0 revolutions are pushing the spread of collaborative robots (cobots) [1]. New workspaces where automated machines cooperate with humans are rising quickly but, from a research point of view, the experience of a worker interacting with a cobot still remains largely unexplored. For these reasons, a great research effort in social capabilities of robots is required to keep up with the changes of today's industrial environments and to understand how human-robot interactions (HRI) can resemble a social experience similar to an everyday human-human work interaction. This brings with it the promise of reducing social isolation and increasing well-being at the workplace. In fact, robust evidence indicates that social connections provide great benefits both physically and cognitively, as reported in [2].

In these regards, the analysis of human-human interactions immediately highlights the importance for verbal and non-verbal communication. In every socially interactive scenario, motor correlates such as lip-syncing, head nods, deictic gestures and gaze movements are abundant and play a great role in expressing emotions and intentions and clarifying unexpressed details laying the ground for the actual content of the communication [3]. Also, gestures can be considered more accurately as a complementary channel of communication. Research is focusing both on the generation and on the combination of such actions [4], [5]. It is clear that robots, and especially industrial cobots, do not offer any of these capabilities that are fundamental to build a natural and social interaction. From a conceptual point of view, a virtual avatar could act as a mediator between a cobot and the operator, promoting a more natural and social experience with what is often considered just a tool. In fact, software agents on a screen can easily move in lifelike ways and reproduce sets of actions that are impossible for today's industrial robots, whereas physical embodiment and presence increases salience and importance of the entity compared to two dimensional entities [6]. Starting from the hypothesis that this last statement can be considered true also for technology, studies demonstrate that physical co-located robots, moving in space and able to manipulate objects, are generally perceived as more anthropomorphic and more engaging [7]. On the basis of the above observations, it appears that the integration of the physical capabilities of a robot with the verbal and non-verbal skills provided by a virtual character may represent a viable solution to enhance the perception of the system as a social entity, that is increase the social presence of the entity, and hence, social engagement [8]. Depending on the perceived social role, this would also influence the interaction and the well-being at the workplace [9].

This study is a first attempt to tackle this challenge. An online pilot study with prerecorded video interactions between a human an two cobot-avatar configurations was set up. The goal is to gain preliminary insights into which features this kind of systems should have to be perceived as an embodied social entity. First-person perspective videos were used in order to facilitate the viewer's immersion in the worker's perspective [10] and to increase learning effects on assembly tasks [11].

## II. BACKGROUND

One way to realize the aforementioned multimodal communication is through the use of tele-operated humanoid robot avatars. For example, in [12] the NAO robot platform is used to directly transmit the tele-operator's actions and to prove that gestures integrated with speech are understood by participants as well as when produced by humans. A common field of application for teleoperated robotics is conferencing and in [13] it was demonstrated that physical embodiment enhances social telepresence. However, teleoperation is an extremely demanding approach since a human operator is always required to control the machine and engage with the user and, therefore, it may not be applicable to all situations. A completely different approach is instead presented in [14], where the authors had the participants experience the body of a humanoid robot as if it was their own with the goal of analyzing aspects related to guilt. Even though the embodiment under analysis in this study is not between the robot and a virtual character but between the cobot and the participant, it is still relevant to know that it is possible for humans to project themselves into machines and therefore to consider them as acceptable social entities. In this regards, this result is an additional confirmation of what started to become clear with the classic animation experiment presented in [15]: humans have a strong tendency to impose narrative even on non-humanoid interactions. Considering humanoid robots, an interesting take on the matter is provided, for instance, by the Baxter robot [16] for which a tablet is used to display the "robot's eyes". However, there are several applications where non-humanoid robots, for which this approach may not be applicable, are often deployed (e.g., collaborative industrial scenarios). Previous work has also shown that immersive virtual environments can facilitate social and emotional learning [17]. A question therefore arises: can a non-humanoid robot be perceived as an embodied social entity? On this topic, an interesting analysis is presented in [18] in which the authors demonstrate that just observing a non-humanoid robot touching objects, such as a curtain, can induce haptic sensations in the viewer.

[1]National Research Council of Italy, Institute of Intelligent Industrial Technologies and Systems for Advanced Manufacturing, Lecco, Italy. matteo.lavit@stiima.cnr.it
[2]Industrial Engineering Department, University of Bologna, Bologna, Italy.
[3]German Research Center for Artificial Intelligence, Saarbrucken, Germany.

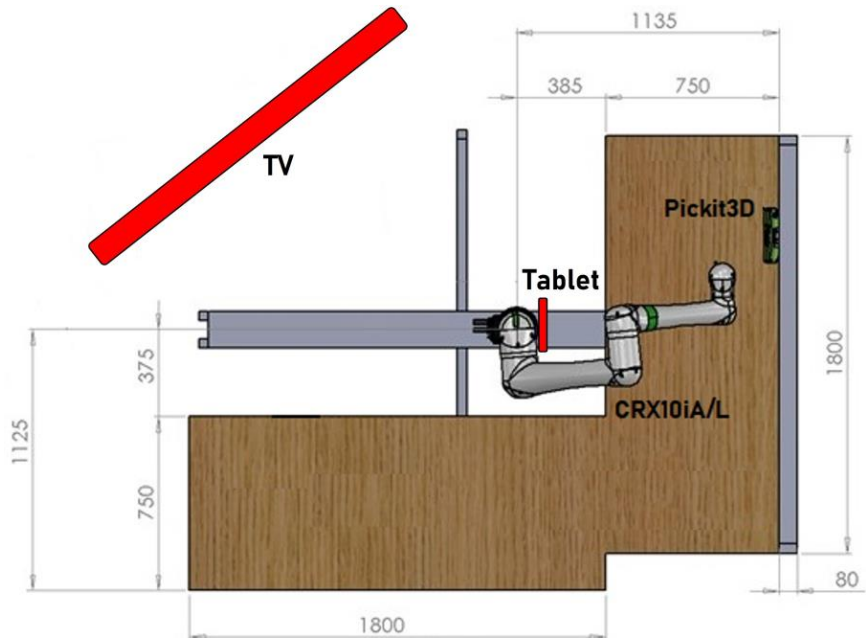

Fig. 1: Overview of the workcell used for video acquisition

These results seem to suggest that humans are able to project themselves even into non-humanoid machines and, therefore, that perceiving these types of machines as embodied social entities is, in fact, possible. However, the results obtained in [18] very much depend on the personality of the participants and on how the experience was proposed, meaning that the hypothesis just presented may not be easily generalized for day-to-day activities. With the aim of facilitating said generalization, the yet unexplored integration of a virtual character with a physical non-humanoid robot may represent a promising approach to embodiment if the two entities are perceived as a whole, meaning that the robot has to be perceived as the physical interface that allows the virtual character to interact with the surroundings. We present first steps into this direction.

## III. MATERIALS AND METHODS

In order to record two videos depicting a worker's interaction with a cobot-avatar integrated system, the industrial production cell represented in Figure 1 was reproduced in a lab-based environment. The experimental workcell has been designed for the collaborative assembly of the 3D-printed planetary gearbox shown in Figure 2. Half of the assembly is carried out by the cobot itself while the worker has responsibility for the other half. Direct collaboration between the two takes place when the two subassemblies are finally put together, with the cobot keeping its part in a precise orientation and the user performing the meshing of the gears.

For this purpose, as depicted in Figure 1, a Fanuc CRX10iA/L [19] collaborative robot is rigidly connected to an L-shaped workbench and equipped with a Pickit3D camera [20] for parts detection. The right side of the table is where all the parts needed by the cobot are stored and where the cobot performs its own part of the assembly. The table on the left, instead, is used by the worker for both the subassembly task and the final collaborative assembly step. Moreover, the worker has to restock the cobot with a buffer of components. The detection camera allows the cobot to look for the required component, which is then grabbed using a Robotiq Hand-E gripper [21] and used for assembly purposes. Always with reference to Figure 1, two configurations of cobot-avatar were designed to capture aspects of embodiment through perceived Unity, Social Presence, and role: either a large TV screen is positioned in the corner of the workcell, in order to be always visible by the user, or a tablet is strapped directly on the basis of the cobot in order to move together with it and always face in the direction of the gripper. From a software point of view, the whole system is driven within ROS [22] and directly integrated with Visual SceneMaker [23] for task definition and synchronization. In particular, the avatar visualizer module was developed using the YALLAH framework [24], [25] that allows direct customization using Blender 3D [26] and deployment as a stand-alone Unity3D [27] application. A screenshot of the virtual character used for this study is reported in Figure 3.

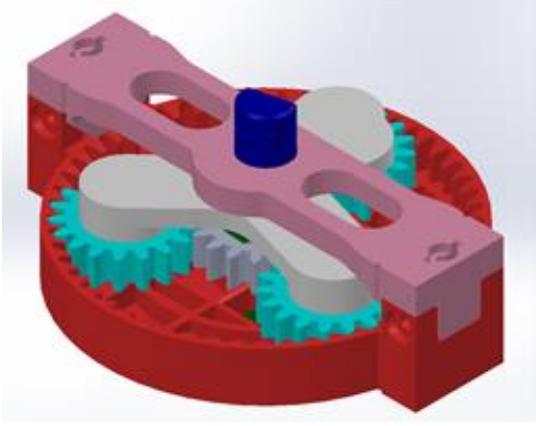

Fig. 2: 3D model of the finished product

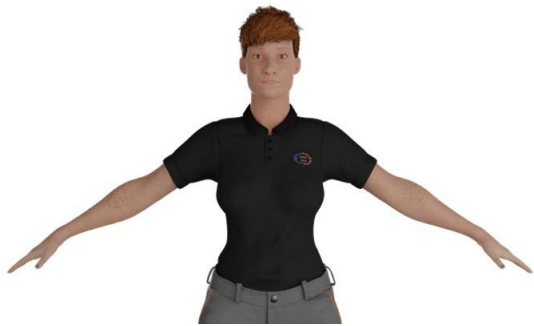

Fig. 3: Screenshot of the virtual character

For each of the proposed configurations, a video showing the worker's interaction with the cobot-avatar system was recorded using a standard camera phone and integrated in the online survey administered using SoSci-Survey [28].

## IV. CASE STUDY

As previously mentioned, the use case under analysis in this study is the collaborative assembly of a 3D-printed planetary gearbox. For this specific task, two main collaboration sessions can be identified:

- The user restocks the cobot table with components while the cobot is working on its part of the subassembly.
- The cobot holds its subassembly in a precise orientation while the user assembles it with the other subassembly by correctly meshing the gears.

In order to present both interactions, first-perspective videos were shown, composed of two subsequent scenes, put together with a cross-fade to show that some time may have passed between the two events. Below, the script for the first interaction session is reported:

> WORKER: Good morning! [*Worker speaks to the resting system.*]
> SYSTEM: Good morning! [*Cobot wakes up and avatar waves at the worker.*]
> WORKER: Let's get to work.
> SYSTEM: OK. [*Avatar looks at the component and cobot moves to pick it.*]
> WORKER: I have some parts for you. [*Worker looks at his hand full of components and places them on the table.*]
> WORKER: OK. [*Worker moves towards his side of the workcell.*]
> SYSTEM: Thanks!

Likewise, the script for the second interaction session is reported below:

> WORKER: Hey, I'm almost done. [*Worker is handling some parts and looks at the system.*]
> SYSTEM: Here I am. [*Cobot brings the finished subassembly in front of the user, the user completes the assembly and retrieves the product.*]
> SYSTEM: Thanks!
> WORKER: Great! Thank you! [*Cobot moves back to its table while the avatar is looking in that direction. The worker puts the finished product in a box.*]

The main goal of this pilot study was to gain insights into two different configurations of the cobot-avatar integrated system: the TV screen in the corner and a tablet strapped at the base of the cobot itself. The main difference between the two proposed setups is the selection of the screen where the avatar will be displayed, as can be seen in Figure 4. Both configurations were conceived to create the perception of an embodied social agent and Social Presence [6], [13]. Therefore the two videos were scripted with the aim of reporting the same behaviours for all the entities in play (robot, virtual agent, human worker). It must be noticed that the scripts have been designed in order not to suggest or force the recognition of Unity on the viewer, for instance by having the virtual character speak about the robot as if it was part of its body. Despite the attempt to have the two configurations as similar as possible, some differences can still be identified. The tablet setup could lead to a stronger perception of Unity, and hence embodiment, between cobot and avatar, since the two units are co-located and move synchronously inside the cell, and, hence, increase Social Presence [13]. Because the Avatar is not constantly looking at the worker, this setup may be perceived rather as a coworker and, therefore, the Social Role of a colleague may be ascribed. On the contrary, the TV setup, allows the avatar to always be visible to the user during the task and hence Social Presence may perceived on account of the avatar alone. However, due to the spatial distance, and the lack of synchronised motion, perceived Unity might be lower. Moreover, since the Avatar always faces the worker this could lead to the feeling of being supervised. These differences, give us the possibility to test the differential relation between Social Presence, Unity and Social Role in embodied cobot-agent configurations. We therefore hypothesise that:

1) *The Tablet configuration will increase the perception of the colleague Social Role, of Social Presence and of Unity more than in the tv configuration.*
2) *The TV configuration will increase the perception of the supervisor Social Role more than in the Tablet configuration.*
3) *Perceived Social Presence will positively correlate with perceived Unity, and Social Role.*

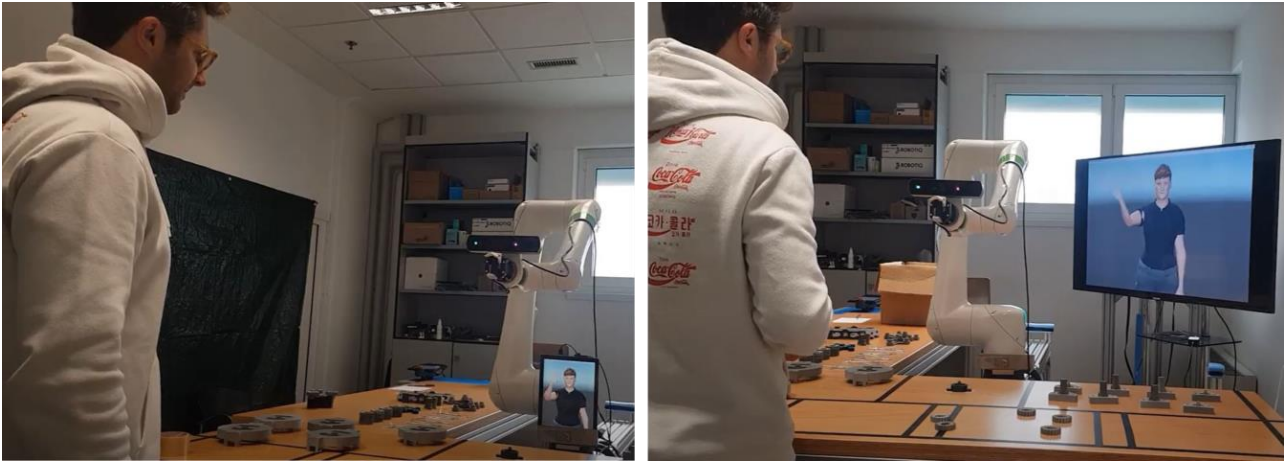

Fig. 4: On the left the tablet configuration, on the right the TV configuration

## V. ONLINE QUESTIONNAIRE

Two questionnaire in total, one for each condition/set-up, were prepared. The questionnaire contained one video and a set of items, consisting of four scales: The Social Role Scale (Collegiality and Supervision), the Social Presence Scale and the Unity Scale, as well as items concerning consent and the ability to imagine oneself in the situation. For all scales a 5-point Likert Scale, ranging from Strongly Disagree to Strongly Agree, was used. Both the Collegiality Scale and the Supervision Scale related to the relationship of the worker to the system. The Social Presence Scale relates to the feeling of working with someone, rather than something. Lastly, the Unity Scale measures whether the avatar and the Cobot are seen as one, and not as two independent systems.

The Social Presence Scale is based on the Social Presence Questionnaire of Lin [29]. The formulation of the questions were adapted to fit the setting of the current study. Since we could not find any existing questionnaires for the other scales, the items were self created, based on associated properties. For Unity, we oriented ourselves on the dictionary definition [30] and derived three properties namely: Oneness, Harmony and Relatedness of the Parts. Some example questions include:

- The avatar and the robot are in sync.
- The avatar sees everything the robot sees.
- If the avatar malfunctions the robot would not function properly.

The two Social Roles, were also construed based on common expectations.

The scales were piloted with $N = 10$ participants, mostly undergraduates from Italy and Germany to reduce the number of questions. The Unity Scale was reduced to five items, and the rest to three items. Here is the internal consistency for each scale:

- Colleagiality Scale: Cronbach Alpha = 0.879
- Supervision Scale: Cronbach Alpha = 0.787
- Social Presence Scale: Cronbachs Alpha = 0.771
- Unity Scale: Cronbach Alpha = 0.932

As seen above all values range from 0.7 to 0.95 which we deem to be acceptable [31]. Lastly, we used a control item, to measure how the participants felt about having to imagine themselves in the situation of the worker. This control question asked how easy it was for the participants to imagine themselves in the position of the worker and used a 7-point Likert Scale.

For the online study, participants from Germany and Italy (mainly people working in the building and students) were recruited on a voluntary basis without any compensation and took part between the 5th of May and the 1st of June to. Exactly one week after a participant received the first questionnaire, they received the second one. We distributed the participants randomly to two groups. Group 1 saw the Tablet condition first and the TV condition second, for Group 2 it was the other way around meaning that both configurations were experienced by both groups within the period of one week.

| Hypothesis Testing | | | | | | |
|---|---|---|---|---|---|---|
| Scale | Condition | Test | Correlation | Degrees of Freedom (df) | T Value (t) | Significance (p) (alpha = 0.05) |
| Collegiality | TV vs Tablet | Paired T-Test | | 19 | -0.028 | **0.978** |
| Supervision | TV vs Tablet | Paired T-Test | | 19 | 0.301 | **0.767** |
| Social Presence | TV vs Tablet | Paired T-Test | | 19 | t < 0.001 | **1** |
| Unity | TV vs Tablet | Paired T-Test | | 19 | 0.946 | **0.356** |
| Social Presence & Collegiality | TV & Tablet | Pearson Product-Moment Correlation | .74 | 38 | | **p < 0.001\*\*** |
| Social Presence & Supervision | TV & Tablet | Pearson Product-Moment Correlation | -.23 | 38 | | **0.154** |
| Social Presence & Unity | TV & Tablet | Pearson Product-Moment Correlation | .65 | 38 | | **p < 0.001\*\*** |

TABLE I: Results

## VI. RESULTS

After collecting the data, we excluded all trials who did not fulfill following conditions:

1) The participant has given consent to their data being used.
2) The participant has finished both questionnaires once.

After cleaning the data we were left with n = 20 participants. The scales were calculated as the means of the respective items. The data of all scales were distributed normally (p-values > .05). Due to the small sample of the pilot study, we did not use a multivariate analysis and just tested the directed hypotheses with paired t-test to compare for each tested outcome, and Pearson Product-Moment Correlation to test the expected correlations.

The control imagination question showed moderate immersion ($M = 3.32$, $Median = 3$, $SD = 1.82$). When comparing the overall means of the two groups, we did not find any significant differences (Group 1: $M = 3.198$, $Median = 3.333$, $SD = 0.869$; Group 2: $M = 3.183$, $Median = 3.333$, $SD = 0.862$). The overall mean answers in all scales were also very similar and the groups answer to a scale of one condition never differed by more than 0.5. Also both groups scored the Tablet setup slightly, but not significantly, higher (for Group 1: $M = 0.132$ and for Group 2: $M = 0.013$), which leads us to believe that the order of viewing had no effect on the ratings. Accordingly, the results show no significant difference concerning the two configurations (see Table I). We found a significant correlation between Social Presence, Collegiality and Unity, but no significant correlation between Social Presence and Supervision. There was also no correlation between Supervision and Collegiality.

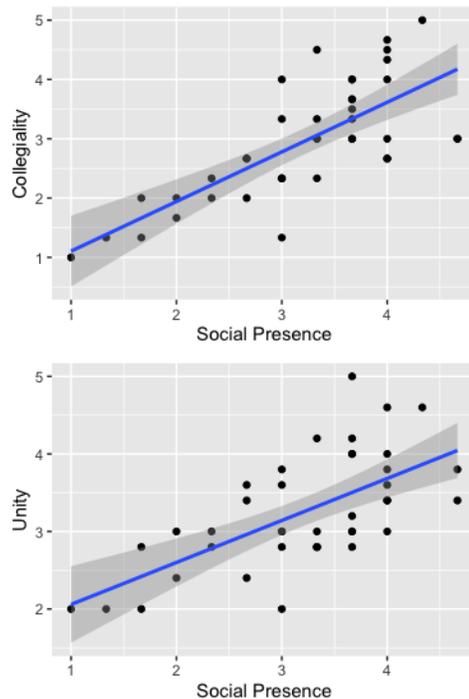

Fig. 5: Scatter plots of the correlation of Social Presence - Collegiality and Social Presence - Unity. The regression line uses the least sums of squares method and the grey area represents the 95% CI of the best-fit line.

## VII. DISCUSSION AND CONCLUSION

The lack of difference between the conditions regarding Social Presence and Unity, was unexpected. It seems like the avatar alone was enough to raise the Social Presence in the TV condition, and thus Unity as well [13]. This means that both configurations can successfully create perceptions of embodiment and Social Presence [6], [13]. The confirmed correlation between Social Presence, Unity and Collegiality, and the lack of correlation between Social Presence and Supervision point to an interest direction for future research, which can test the direction of the effect with regard to the influence of Social Role and Social Presence. The implication for designing embodied social cobots is rather that a colleague-avatar would be more conducive to supporting Social Presence. However, the effect has to be tested also in connection to the behaviour of the avatar. The small sample of a pilot and the moderate degree of imagining oneself in the situation, as measured by the treatment check indicate, as expected, the need for larger scale in-presence studies, to be able to verify the results, in alignment with previous work on the benefits of co-located agents [7] and on the accuracy of predictions based on imagining future events [32]. Future work is therefore needed to collect a much larger in-person database since this approach would provide more robust results, independent from the way the videos were recorded. Also, the authors intend to explore additional behaviours of both the cobot and the virtual character that may be helpful in enhancing the perceived social presence.

## VIII. ACKNOWLEDGMENT

This project has received funding from the European Union's Horizon 2020 research and innovation programme under grant agreement No 847926.

## REFERENCES

[1] L. Liu, F. Guo, Z. Zou, and V. Duffy, "Application, development and future opportunities of collaborative robots (cobots) in manufacturing: A literature review," *International Journal of Human-Computer Interaction*, 2022, cited By 0.
[2] J. Holt-Lunstad, "Fostering social connection in the workplace," *American Journal of Health Promotion*, vol. 32, no. 5, pp. 1307–1312, 2018.
[3] N. Mavridis, "A review of verbal and non-verbal human-robot interactive communication," *Robotics and Autonomous Systems*, vol. 63, no. P1, pp. 22–35, 2015, cited By 226.
[4] J. Cassell, T. Bickmore, M. Billinghurst, L. Campbell, K. Chang, H. Vilhjailmsson, and H. Yan, "Embodiment in conversational interfaces: Rea," 1999, pp. 520–527, cited By 334.
[5] J. Cassell, H. Vilhjálmsson, and T. Bickmore, "Beat: The behavior expression animation toolkit," 2001, pp. 477–486, cited By 89.
[6] H. Kawamichi, Y. Kikuchi, and S. Ueno, "Magnetoencephalographic measurement during two types of mental rotations of three-dimensional objects," *IEEE Transactions on Magnetics*, vol. 41, no. 10, pp. 4200–4202, 2005, cited By 6.
[7] S. Kiesler, A. Powers, S. Fussell, and C. Torrey, "Anthropomorphic interactions with a robot and robot-like agent," *Social Cognition*, vol. 26, no. 2, pp. 169–181, 2008, cited By 190.
[8] K. Kreijns, K. Xu, and J. Weidlich, *Social Presence: Conceptualization and Measurement*. Educational Psychology Review, 2022, vol. 34, no. 1.
[9] C. Ray, F. Mondada, and R. Siegwart, "What do people expect from robots?" in *IEEE/RSJ International Conference on Intelligent Robots and Systems*, vol. 41, no. 1, 2008, pp. 22–26.
[10] J. Fukuta and J. Morgan, "First-person perspective video to enhance simulation," *Clinical Teacher*, vol. 15, no. 3, pp. 231–235, 2018.
[11] L. Fiorella, T. van Gog, V. Hoogerheide, and R. E. Mayer, "It's all a matter of perspective: Viewing first-person video modeling examples promotes learning of an assembly task," *Journal of Educational Psychology*, vol. 109, p. 653–665, 2017.
[12] P. Bremner and U. Leonards, "Iconic gestures for robot avatars, recognition and integration with speech," *Frontiers in Psychology*, vol. 7, 2016, cited By 23.
[13] K. Tanaka, H. Nakanishi, and H. Ishiguro, "Comparing video, avatar, and robot mediated communication: Pros and cons of embodiment," in *Collaboration Technologies and Social Computing*, T. Yuizono, G. Zurita, N. Baloian, T. Inoue, and H. Ogata, Eds. Berlin, Heidelberg: Springer Berlin Heidelberg, 2014, pp. 96–110.
[14] L. Aymerich-Franch, S. Kishore, and M. Slater, "When your robot avatar misbehaves you are likely to apologize: An exploration of guilt during robot embodiment," *International Journal of Social Robotics*, vol. 12, no. 1, pp. 217–226, 2020, cited By 6.
[15] F. Wick, A. Soce, S. Garg, R. Grace, and J. Wolfe, "Perception in dynamic scenes: What is your heider capacity?" *Journal of Experimental Psychology: General*, vol. 148, no. 2, pp. 252–271, 2019.
[16] C. Fitzgerald, "Developing baxter," in *2013 IEEE Conference on Technologies for Practical Robot Applications (TePRA)*, 2013, pp. 1–6.
[17] M. C. C. Tan, S. Y. L. Chye, and K. S. M. Teng, ""In the shoes of another": immersive technology for social and emotional learning," *Education and Information Technologies*, pp. 1–24, 2022.
[18] L. Aymerich-Franch, D. Petit, G. Ganesh, and A. Kheddar, "Object touch by a humanoid robot avatar induces haptic sensation in the real hand," *Journal of Computer-Mediated Communication*, vol. 22, no. 4, pp. 215–230, 2017, cited By 12.
[19] Fanuc crx10ia/l. [Online]. Available: https://www.fanuc.eu/it/en/robots/robot-filter-page/collaborative-robots/crx-10ial
[20] Pickit3d camera. [Online]. Available: https://www.pickit3d.com/en/
[21] Robotiq hand-e gripper. [Online]. Available: https://robotiq.com/products/hand-e-adaptive-robot-gripper
[22] M. Quigley, K. Conley, B. P. Gerkey, J. Faust, T. Foote, J. Leibs, R. Wheeler, and A. Y. Ng, "Ros: an open-source robot operating system," in *ICRA Workshop on Open Source Software*, 2009.
[23] P. Gebhard, G. Mehlmann, and M. Kipp, "Visual scenemaker-a tool for authoring interactive virtual characters," *Journal on Multimodal User Interfaces*, vol. 6, no. 1-2, pp. 3–11, 2012, cited By 42.
[24] F. Nunnari and A. Heloir, "Yet another low-level agent handler," *Computer Animation and Virtual Worlds*, vol. 30, no. 3-4, p. e1891, 2019, e1891 cav.1891. [Online]. Available: https://onlinelibrary.wiley.com/doi/abs/10.1002/cav.1891
[25] (2021) Yallah - yet another low-level agent handler. [Online]. Available: https://github.com/yallah-team/YALLAH/
[26] Blender. [Online]. Available: https://www.blender.org/
[27] Unity. [Online]. Available: https://unity.com/
[28] Sosci-survey. [Online]. Available: https://www.soscisurvey.de/
[29] G.-Y. Lin, "Social presence questionnaire of online collaborative learning: Development and validity," 2005.
[30] Merriam-Webster, "Unity." [Online]. Available: https://www.merriam-webster.com/dictionary/unity
[31] T. M and D. R., "Making sense of cronbach's alpha," *International journal of medical education*, vol. 2, pp. 53–55, 2011.
[32] M. Hoerger, S. W. Quirk, R. E. Lucas, and T. H. Carr, "Cognitive determinants of affective forecasting errors," *Judgment and Decision Making*, vol. 5, no. 5, pp. 365–373, 2010.